\documentclass[11pt,a4paper]{article}
\usepackage[affil-it]{authblk}
\usepackage[hyperref]{acl2020}
\usepackage{times}
\usepackage{soul}
\usepackage{xcolor}
\usepackage{latexsym}
\usepackage{graphicx}
\usepackage{booktabs}
\usepackage{multirow}
\usepackage{placeins}
\usepackage{tablefootnote}

\usepackage[colorinlistoftodos,prependcaption,textsize=tiny]{todonotes}
\newcommand{\clarify}[1]{\textcolor{red}{\textbf{Do we want to keep this?}}}

\usepackage{color}
\definecolor{blue}{RGB}{0, 93, 170}			
\definecolor{darkgreen}{RGB}{0, 102, 0}

\usepackage{microtype}

\usepackage{array}
\usepackage{etoolbox}
\usepackage{url}

\hypersetup{breaklinks=true}

\newcounter{rowcntr}[table]
\renewcommand{\therowcntr}{\thetable.\arabic{rowcntr}}

\newcolumntype{N}{>{\refstepcounter{rowcntr}\therowcntr}c}

\AtBeginEnvironment{tabular}{\setcounter{rowcntr}{0}}

\aclfinalcopy

\makeatletter
\newcommand{\printfnsymbol}[1]{%
  \textsuperscript{\@fnsymbol{#1}}%
}
\makeatother
\title{Reading Comprehension as Natural Language Inference:\\ A Semantic Analysis}
\author[1]{Anshuman Mishra \thanks{~~Equal contribution.}~}
\author[1]{Dhruvesh Patel\printfnsymbol{1}} 
\author[1]{Aparna Vijayakumar\printfnsymbol{1}}
\author[1]{\\Xiang Li}
\author[2]{Pavan Kapanipathi}
\author[2]{Kartik Talamadupula}
\affil[1]{ College of Information and Computer Sciences,   University of Massachusetts Amherst}
\affil[2]{IBM Research}

\date{}

\begin{document}
\maketitle
\begin{abstract}
In the recent past, Natural language Inference (NLI) has gained significant attention, particularly given its promise for downstream NLP tasks. 
However, its true impact is limited and has not been well studied. 
Therefore, in this paper, we explore the utility of NLI for one of the most prominent downstream tasks, viz. Question Answering (QA).  
We transform the one of the largest available MRC dataset (RACE) to an NLI form, and compare the performances of a state-of-the-art model (RoBERTa) on both these forms. We propose new characterizations of questions, and evaluate the performance of QA and NLI models on these categories.
We highlight clear categories for which the model is able to perform better when the data is presented in a coherent entailment form, and a structured question-answer concatenation form, respectively.
\end{abstract}

\section{Introduction}
Given two sentences, a premise and a hypothesis, the task of Natural Language Inference (NLI) is to determine whether the premise entails the hypothesis or not. \footnote[2]{The ``not entailment'' can further be subdivided into ``neutral'' and ``contradiction''. However, we only use the two-class version of the problem in this work.} 
The concept of semantic entailment is central to natural language understanding \cite{van2008brief, maccartney2009natural} and therefore,  NLI models have been used to help with various downstream tasks like reading comprehension \cite{Trivedi2019},  summarization \cite{falke2019ranking,kryscinski2019evaluating}, and dialog systems \cite{dnli}. However, the performance of an NLI system on these downstream tasks has not been studied with respect to semantic or reasoning categories.

In this work, we use NLI to perform the task of multiple choice reading comprehension (MRC, or RC). We analyse the performance of an NLI model on this task through the lens of semantics by identifying the reasoning categories (type of questions) where it is beneficial to use an NLI model.

Drawing inspiration from the prior work in the area \cite{clark2018think, Demszky2018, Trivedi2019}, we use rule-based conversion to create an NLI version of the largest available RC dataset - RACE \cite{lai2017race}. We train a RoBERTa based RC model on the original dataset, and a similar RoBERTa based NLI model on the NLI version of the dataset. We evaluate and analyse the performance of both these models by characterizing the question types that are better suited for an NLI model and a QA model.

\section{Related Work}
\begin{figure*}
\centering
\begin{tabular}{cc}
  \includegraphics[width=0.4\linewidth, height=7cm]{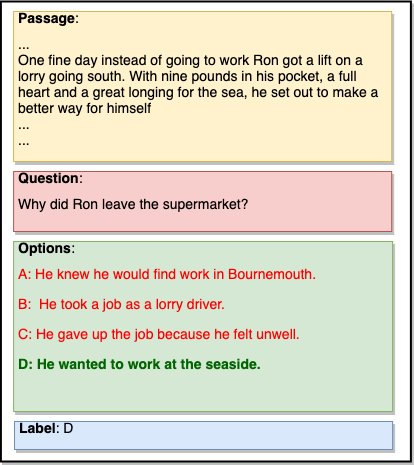} \hspace{5mm}&   \includegraphics[width=0.4\linewidth, height=7cm]{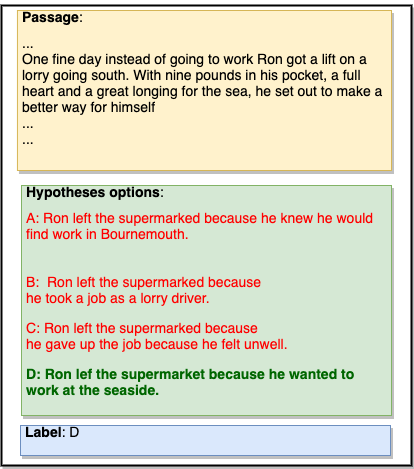} \\
(a) QA sample & (b) Converted NLI sample \\[6pt]
\end{tabular}
\vspace{-4mm}
\caption{A RC sample with multiple answer choices converted to an NLI sample.}
\label{fig:conversion}
\end{figure*}

Reading comprehension (RC) is one of many potential downstream tasks that can benefit from NLI \cite{maccartney2009natural}. It is easy to see that RC naturally reduces to a two-class NLI problem; specifically, it can be cast as the task of identifying if a given piece of text entails the statement formed by converting a question and a potential answer to an assertive statement (hypothesis).

Given the intuitive conversion between RC and NLI, \citet{Demszky2018} designed both a rule-based conversion system as well as a trained neural model to convert question-answering datasets such as SQuAD \cite{rajpurkar2016squad} and RACE \cite{lai2017race} to an NLI form. However, it is unclear what these converted NLI datasets  offer compared to the original question-answering datasets w.r.t semantics. We show that converting a RC task to an NLI task helps in answering certain types of questions. This establishes the usefulness of the converted datasets. 

\citet{jin2019mmm} show that despite the different form of NLI and QA tasks, performing coarse pretraining of models on NLI datasets like SNLI \cite{Bowman2015} and MultiNLI \cite{mnli} not only improves the performance of these models on downstream reading comprehension tasks, but also helps with faster convergence. We show that -- for certain types of questions in reading comprehension datasets -- simply transforming the task to NLI can show improvement in performance, even without pretraining on any NLI dataset.

\citet{Trivedi2019} introduced a learnt weight-and-combine architecture to effectively re-purpose pretrained entailment models (trained on SNLI and MultiNLI) to solve the task of multi-hop reading comprehension. 
They show that for certain datasets, this strategy can produce good results. 
However, their study focuses mainly on improving model performance using a pre-trained NLI model, and lacks an analysis of the reasoning differences arising due to the different form of the NLI and QA tasks. We focus our analysis on this aspect.


\section{NLI for Reading Comprehension}

This section describes our experimental setup for comparing a QA based approach and an NLI based approach for the task of reading comprehension. We first obtain a parallel NLI and QA dataset by converting existing RC dataset into an NLI dataset. We then train two models, one on each form of the data, and analyse their performance.

\subsection{Converting RC to NLI}
\label{sec:conversion}


We use the RACE dataset \cite{lai2017race} for our experiments. It is a large-scale reading comprehension dataset comprising of questions collected from the English exams for junior Chinese students. Each question contains four answer options, out of which only one is correct. 
However, about 44\% of the RACE dataset consists of cloze style (fill-in-the-blank) questions which are already in NLI form. Hence, in order to have a fair comparison, we only use the subset of RACE dataset which does not contain cloze style questions. This subset consists of $48890$ train, $2496$ validation and $2571$ test examples.

We convert a RC example into an NLI example by reusing the passage as premise and paraphrasing the question along with each answer option as individual hypotheses as shown in Figure \ref{fig:conversion}. Specifically, we generate the dependency parse of both the question and the answer option by using Stanford NLP package \cite{stanfordnlp}, then we follow the conversion rules proposed in \citet{Demszky2018} to generate a hypothesis sentence\footnote{Appendix \ref{app:conversion} presents example conversions generated using these rules.}. We make a few additions to these rules to handle a some peculiar question categories in the RACE dataset. The most prominent of the added rules is the one for questions containing \textit{``which of the following are (not) true''}. Such questions are very frequent (about 6\% of all questions) and are not handled correctly by the rules in \citet{Demszky2018}.

\subsection{Model}
\label{sec:model}

In order to perform apple-to-apple comparison we use the same model architecture for both QA and NLI. Specifically, we use the state-of-the-art reading comprehension model -- consisting of a RoBERTa model (pretrained on the masked language modeling objective) as the encoder and a two layer feed-forward network on its [CLS] token as the classification head -- as described in \citet{liu2019roberta}. The input sentence is the combination of the passage and its hypothesis. The hypothesis are created using the rule-based conversion method mentioned in Section \ref{sec:conversion} (NLI setup) or by concatenation of question and answer option (QA setup).

\section{Analysis}
\label{sec:quantitative}
\begin{table}[]
\centering
\begin{tabular}{@{}lllll@{}}
\toprule
\multirow{2}{*}{Dataset} & \multicolumn{2}{c}{Dataset Format}          \\ \cmidrule(l){2-3} 
                         & QA & NLI \\ \toprule
RACE                     & \textbf{85.78}  & -                   \\
RACE-subset              & 79.84   & \textbf{82.09}       \\\bottomrule
\end{tabular}
\caption{Accuracy on the test set obtained by using different formats of the data.}
\label{tab:all-results}
\end{table}


Table \ref{tab:all-results} shows the accuracy achieved by the RoBERTa model on the RACE dataset and its subset when presented in different forms. \label{subsec:quantitative_race} As we can see, the NLI model performs much better than the QA model on RACE subset.
We think that the reason for this is the more natural form of the hypothesis statement used by the NLI model compared to the Q+A concatenation form used by the QA model.
Moreover, while the RACE-subset consists of only those questions which have question-words such as \{\textit{who, what, when...}\}, about $95\%$ of the rest of the dataset, i.e. RACE $\setminus$ RACE-subset, consists of only fill-in-the-blank (FITB) type questions. These FITB questions are largely with the blank at the end of the question and a naive question-answer concatenation is very similar to a NLI form hypothesis. We believe that helps the QA model to perform better on the full RACE dataset compared to the RACE-subset, where NLI model is able to outperform the QA model showing the clear benefits of coherent conversion on complex question formulations (such as \textit{W} word questions).

In order to analyse the performance difference from a semantic perspective, we characterize question into $7$ semantic categories by identifying the kind of reasoning required to answer the question. Table \ref{tab:reasoning_categories} succinctly describes the reasoning categories.

\subsection{Categories based on manual analysis}
\label{sec:qualitative}

In order to perform manual analysis, we construct a \emph{delta subset} consisting the $328$ dev set examples on which the predictions of the QA and NLI models differ.
We further divide the \emph{delta subset} into \emph{gain} and \emph{loss} subsets. The \emph{gain subset} consists of questions which the NLI model gets right, but the QA model gets wrong, and the \emph{loss subset} is its complement in the \emph{delta}. 

\begin{table*}[!ht]
\begin{tabular}{p{1.8cm}|p{5.2cm}|p{8.6cm}}
\toprule
\textbf{Category} & \textbf{Description} & \textbf{Example} \\ \midrule
Linguistic Matching & Matching words between the question and a sentence in the passage & \textbf{Passage} : Food cooks quickly in parabolic cookers \\
 &  & \textbf{Question}: If you want to cook food quickly, which kind of sun-cooker is your best choice? \\ \midrule
Main Idea & Require topicality judgements & What's the best title for this passage? \\ \midrule
Negation & Picking the incorrect statement & Which of the following statements is NOT true? \\ \midrule
Dialogue & Can be inferred from a dialogue or direct speech in the passage & By saying "her pen dared travel where her eyes would not", the writer means \\ \midrule
Math & Mathematically combining facts & How many functions of snow are discussed in the text? \\ \midrule
Deductive & None of the above but can be answered precisely from the text & Which of the following statements is TRUE? \\ \midrule
Inductive & None of the above and cannot be answered precisely from the text & How old is most likely the writer's father? \\ \bottomrule
\end{tabular}
\caption{Reasoning Categories (exclusive)}
\label{tab:reasoning_categories}
\end{table*}

\begin{figure}[!ht]
\includegraphics[width=0.45\textwidth]{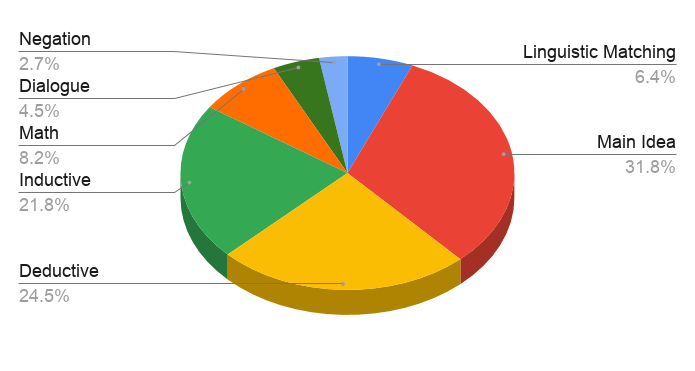}
\caption{Reasoning categories of the gain region}
\label{fig:gain_dist}
\end{figure}

\begin{figure}[!ht]
    \includegraphics[width=0.45\textwidth]{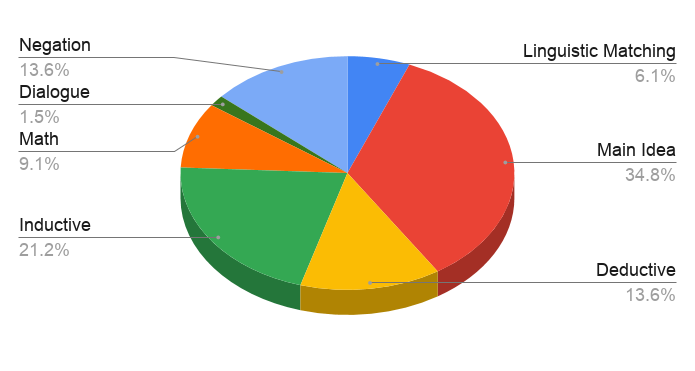}
    \caption{Reasoning categories of the loss region}
    \label{fig:loss_dist}
\end{figure}

We manually annotate these $328$ ($192$ in \emph{gain} and $136$ in \emph{loss}) examples into one of the $7$ categories. However, about half the examples in the \emph{delta subset} were not properly converted by the rules leading to unnatural or incoherent hypothesis sentences. Hence, for the purpose of illustration, we removed these examples leaving a total of $175$ examples ($109$ in \emph{gain} and $66$ in \emph{loss}) to do further analysis. Figures \ref{fig:gain_dist} and \ref{fig:loss_dist} show the distribution of labels over the Gain and Loss regions respectively. The distribution reflects that the QA models clearly outperforms the NLI model in negation questions whereas the NLI model outperforms the QA model in dialogue and deductive reasoning categories.

\begin{table*}[]
    \centering
    \begin{tabular}{p{0.15\textwidth}|p{0.85\textwidth}}
        \toprule
        \textbf{Type} & \textbf{Heuristics} \\
        \bottomrule
        Main Idea & Questions containing the words 'mainly', 'title', 'purpose' or 'topic'\\\hline
        Negation & Questions containing the 'not', 'except' or 'which of the following is wrong'\\\hline
        Dialogue & Passages containing more than 10 quotation marks (")\\\hline
        Math & Questions containing the words 'how many', 'how old' or 'how much'\\\hline
        Deductive & Questions containing the word 'true'\\
        \bottomrule
    \end{tabular}
    \caption{Heuristically Determined Question Types in RACE-subset (non-exclusive).}
    \label{tab:heuristic_categories}
\end{table*}

\subsection{Categories based on heuristics}
We also define another set of non-exclusive categories using heuristics, as described in Table \ref{tab:heuristic_categories}. 
 As shown in Table \ref{tab:heuristic_categories_perf}, the NLI model outperforms the QA model significantly in the dialogue, math and deductive reasoning categories. 
 This overall trend further emphasizes the benefits of proper hypothesis generation as opposed to question and answer concatenation for the reading comprehension task.

\begin{table}[]
    \centering
    \begin{tabular}{c|c|c|c}
        \toprule
        \textbf{Type} & \textbf{Fraction} & \textbf{QA} & \textbf{NLI} \\
        \bottomrule
        Main Idea & 0.12 & 84.19 & 84.83\\\hline
        Negation & 0.06 & \textbf{80.86} & 77.77 \\\hline
        Dialogue & 0.12 & 80.65 & \textbf{83.60} \\\hline
        Math & 0.03 & 45.00 & \textbf{55.00} \\\hline
        Deductive & 0.04 & 81.91 & \textbf{88.29} \\
        \bottomrule
    \end{tabular}
    \caption{Model performances on heuristically determined question types for RACE-Subset.}
    \label{tab:heuristic_categories_perf}
\end{table}

\section{Conclusion}
\label{sec:conclusion}

There is limited work providing a comprehensive analysis of how NLI can be used for QA.
In our work, we show that NLI can be used  for the task of reading comprehension simply by converting the data into NLI form. We convert a large RC dataset into NLI form and perform a comparative study of the performance of the RoBERTa model trained on QA and NLI settings. We propose a categorization of questions that allows for effective comparison of models trained on NLI and QA forms of data. Our analysis clearly shows that using the NLI-based approach is at par with a QA-based approach for most reasoning categories, and it is even better for some. Specifically, we find that questions involving deductive reasoning, dialogue interpretation and math are better handled by a model trained on the NLI form of data than the QA form. However, questions involving negation favor the QA form. Our work allows for careful selection of modeling strategy based on the type of data at hand.





\FloatBarrier
\bibliography{bibliography.bib}
\bibliographystyle{acl_natbib}

\FloatBarrier
\appendix
%

\newpage
\section{Model Architecture}
\label{app:model_arc}

\begin{figure*}
    \centering
    \includegraphics[width=0.7\textwidth]{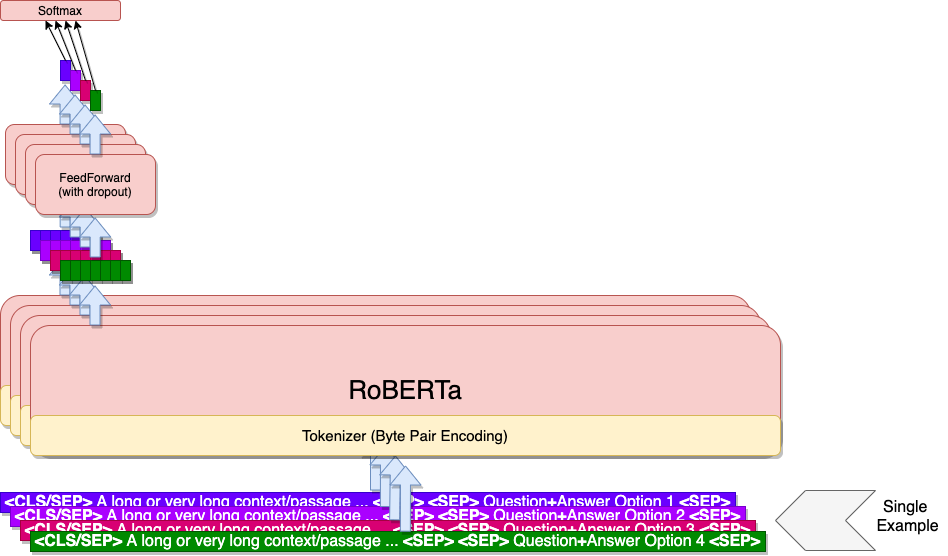}
    \caption{QA model}
    \label{fig:model_qa}
\end{figure*}

\begin{figure*}
    \centering
    \includegraphics[width=0.7\textwidth]{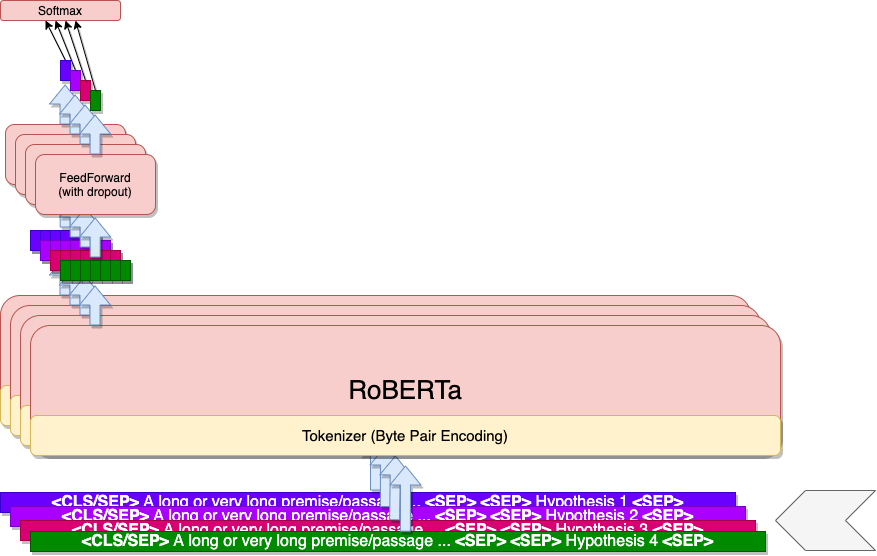}
    \caption{NLI model}
    \label{fig:model_nli}
\end{figure*}

Figures \ref{fig:model_qa} and \ref{fig:model_nli} show the model architecture for the QA and NLI models, respectively. As seen the architecture of the model is the same, the only difference is in the input form.

\section{Hyperparameter Settings}
\label{app:hyperparams}
\begin{table}[!ht]
    \centering
    \begin{tabular}{l|c c}
        \multirow{2}{*}{\textbf{Hyperparam}} 
        & \multicolumn{2}{c}{\textbf{RACE-subset}} \\
        & \textbf{NLI-form} & \textbf{QA-form}  \\
        \bottomrule
        learning rate & 1e-5 & 1e-5  \\ \hline
        weight decay & 0.01 & 0.01 \\ \hline
        warmup steps & 1300 & 1300 \\ \hline
        batch size & 16 & 16 \\ \hline
        max epochs & 4 & 4 \\
        \bottomrule
    \end{tabular}
    \caption{Hyperparamter Setting}
    \label{tab:hyperparams}
\end{table}

Table \ref{tab:hyperparams} lists the hyperparameter settings for both versions of the dataset.
\section{Conversion examples}
\label{app:conversion}
Table \ref{tab:example_conversions_race} shows examples of NLI-form obtained applying rule-based conversion on QA examples from the RACE-subset.

\begin{table*}[ht!]
    \centering
    \begin{tabular}{p{0.3 \textwidth}|p{0.3\textwidth}}
        \textbf{QA example} & \textbf{NLI-form}  \\ \bottomrule
        \textbf{Q:} How do suburban commuters travel to and from the city in Copenhagen at present?
            \break \textbf{A:} About one third of the suburban commuters travel by bike.
            & Suburban commuters travel to about one third of the suburban commuters travel by bike and from the city in Copenhagen at present.
            \\
        \hline
        \textbf{Q:} What's the best title of the passage?
            \break \textbf{A:} Blame! Blame! Blame!
            & The best title of the passage's blame. 
             \\
        \hline
        \textbf{Q:} What influence did the experiment have on Alexander ?
            \break \textbf{A:} He realized that slowing down his life speed could bring him more content.
            & The experiment had he realized that slowing down his life speed could bring him more content on Alexander.
             \\
        \hline
       \textbf{Q:} Which of the following is TRUE about the report findings?
            \break \textbf{A:} The reading scores among older children have improved.
            & The reading scores among older children have improved is TRUE.
             \\
        \bottomrule
    \end{tabular}
    \caption{Examples of Rule-based  conversion applied to samples from the  RACE-subset. }
    \label{tab:example_conversions_race}
\end{table*}

\end{document}